\documentclass[sigconf,noacm]{acmart}
\renewcommand\footnotetextcopyrightpermission[1]{}
\usepackage{algorithm}
\usepackage{algorithmic}
\usepackage{subcaption}
\usepackage{multirow}
\AtBeginDocument{%
  }

\begin{document}

\title{Efficient Masked Image Compression with Position-Indexed Self-Attention}

\author{Chengjie Dai}
\affiliation{%
  \institution{Zhejiang University}
  \city{Hangzhou}
  \country{China}
}

\author{Tiantian Song}
\affiliation{%
  \institution{University College London}
  \city{London}
  \country{United Kingdom}
}

\author{Hui Tang}
\affiliation{%
  \institution{Zhejiang University}
  \city{Hangzhou}
  \country{China}
}

\author{Fangdong Chen}
\affiliation{%
  \institution{Hikvision Research Institute}
  \city{Hangzhou}
  \country{China}
}

\author{Bowei Yang}
\affiliation{%
  \institution{Zhejiang University}
  \city{Hangzhou}
  \country{China}
}

\author{Guanghua Song}
\affiliation{%
  \institution{Zhejiang University}
  \city{Hangzhou}
  \country{China}
}


\begin{abstract}
In recent years, image compression for high-level vision tasks has attracted considerable attention from researchers. Given that object information in images plays a far more crucial role in downstream tasks than background information, some studies have proposed semantically structuring the bitstream to selectively transmit and reconstruct only the information required by these tasks. However, such methods structure the bitstream after encoding, meaning that the coding process still relies on the entire image, even though much of the encoded information will not be transmitted. This leads to redundant computations. Traditional image compression methods require a two-dimensional image as input, and even if the unimportant regions of the image are set to zero by applying a semantic mask, these regions still participate in subsequent computations as part of the image. To address such limitations, we propose an image compression method based on a position-indexed self-attention mechanism that encodes and decodes only the visible parts of the masked image. Compared to existing semantic-structured compression methods, our approach can significantly reduce computational costs.
\end{abstract}




\keywords{Computer Vision, Image Compression}


\maketitle
\pagestyle{plain}

\section{Introduction}
\label{sec: 1}
What elements of an image are most critical to human visual perception? The answer to this question led to the development of lossy image compression. Compared to color, the human visual system is more sensitive to brightness. Therefore, researchers proposed converting the image from RGB space to YCrCb space to perform color downsampling. Similarly, since humans are more sensitive to low-frequency signals compared to high-frequency signals, the DCT transformation can be used to eliminate some high-frequency signals \cite{Wallace1991}. After decades of progress, lossy image compression has achieved remarkable performance. In addition to traditional hand-crafted compression algorithms \cite{Bellard2015,Taubman2002}, neural network-based learned image compression models have also shown significant potential \cite{He2021,He2022, Zou2022,Liu2023}.

In recent years, the widespread use of computer vision models has driven an increasing demand for image compression tailored to downstream high-level vision tasks such as object detection and instance segmentation. Traditional lossy compression methods designed for the human visual system are no longer ideal for these scenarios. As a result, image and video coding for machine vision has become a significant field of study \cite{Duan2020}. Some approaches propose extracting features at the encoding end and compressing these deep intermediate features, which are then decompressed at the decoding end and directly fed into downstream models for inference \cite{Torfason2018, Feng2022, Chen2023}. Although these methods often maintain downstream task accuracy even at high compression rates, manual verification of detected targets is still required in some cases. Since these methods do not reconstruct the image, they cannot support scenarios that require visual confirmation.

Reflecting on the methods used in traditional image compression, we inevitably consider which elements of an image are most critical for high-level vision tasks. High-level vision tasks focus mainly on the semantic understanding of objects and scenes within an image. From this point of view, semantic information plays the key role in such tasks. SSIC (Semantically structured image compression) \cite{Sun2020} generates a semantically structured bitstream (SSB) by leveraging a pre-built object detection toolbox to identify objects and compressing their rectangular regions individually. Building on SSIC, GIT-SSIC (Semantically structured image compression with group-independent transform) \cite{Feng2023} further divides the image into multiple groups with irregular shapes using a customized group mask. These groups are then compressed independently to form the SSB, enabling selective transmission. At the decoding stage, GIT-SSIC reconstructs only the preserved groups. This yields a partial image that can be used for both machine vision tasks and manual verification, thereby supporting both automated and human-in-the-loop applications.

However, while GIT-SSIC applies a semantic mask to get the information needed by downstream tasks, its encoding strategy still relies on the full image. GIT-SSIC first uses an encoder network to process the entire image and then semantically structures the bitstream for grouped compression. As a result, there is a substantial amount of unused information in the image that is still processed by the encoder, leading to unnecessary consumption of computational resources. The computational cost could be significantly reduced by applying a mask to the input image beforehand and encoding only the visible parts. Currently, most image compression networks are designed to operate on complete two-dimensional images. As such, retaining only the visible regions after masking violates the input requirements of these models. These limitations motivate the design of a new image compression framework capable of efficiently handling masked input.

\begin{figure}[h]
\centering
\includegraphics[width=\columnwidth]{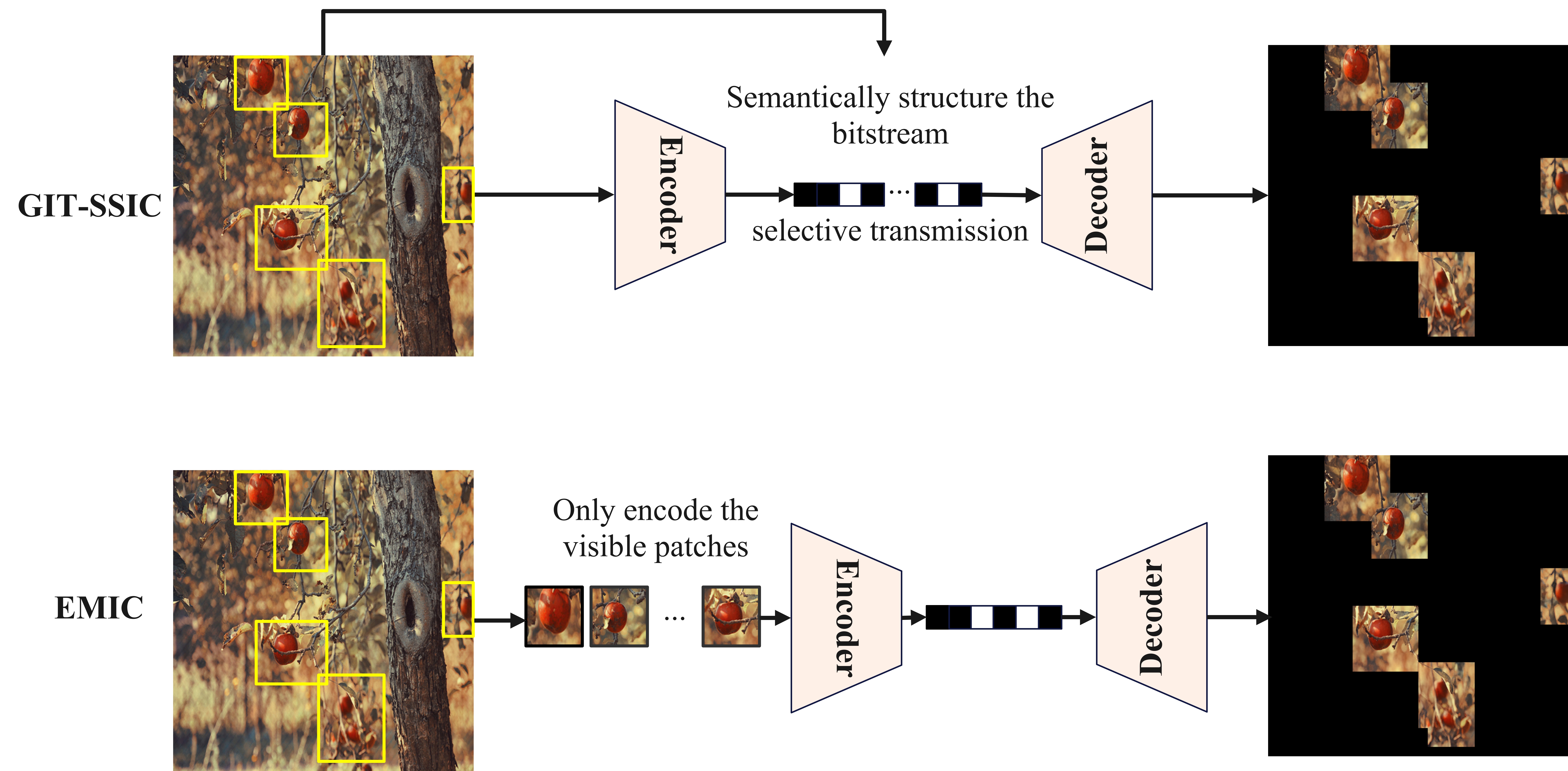}
\caption{Differences between EMIC and GIT-SSIC in Semantic Masking}
\label{fig: semantic_image_compression}
\end{figure}

In this paper, we propose a compression network, EMIC (Efficient Masked Image Compression), which leverages semantic masks to reduce computational costs. The network integrates masked feature extraction and compression within a unified model. Similar to the Vision Transformer (ViT) \cite{Dosovitskiy2020}, the input image is first divided into patches, and information is aggregated across patches after masking via an attention mechanism. A comparison between EMIC and GIT-SSIC is illustrated in Fig. \ref{fig: semantic_image_compression}. EMIC encodes and decodes only the visible patches after masking, thereby avoiding any redundant computation on the masked regions.

Drawing inspiration from RMT \cite{fan2024rmt}, we introduce spatial decay into the attention mechanism using the Manhattan distance. Building on this idea, we propose position-indexed self-attention, which computes Manhattan distances between visible patches using their positional indices during the self-attention process. This approach more effectively incorporates spatial priors among visible patches, leading to improved model performance. The main contributions of our method are summarized as follows:
\begin{itemize}
    \item We propose a novel compression network model that can directly compress the sequence of visible patches formed after the image is masked.
    \item We propose the position-indexed attention method, which use the relative positional relationships of visible patches to enhance the performance of the compression model.
    \item The proposed model significantly reduces FLOPs by performing attention calculations only on visible patches.
\end{itemize}

\begin{figure*}[t]
\centering
\includegraphics[width=0.9\textwidth]{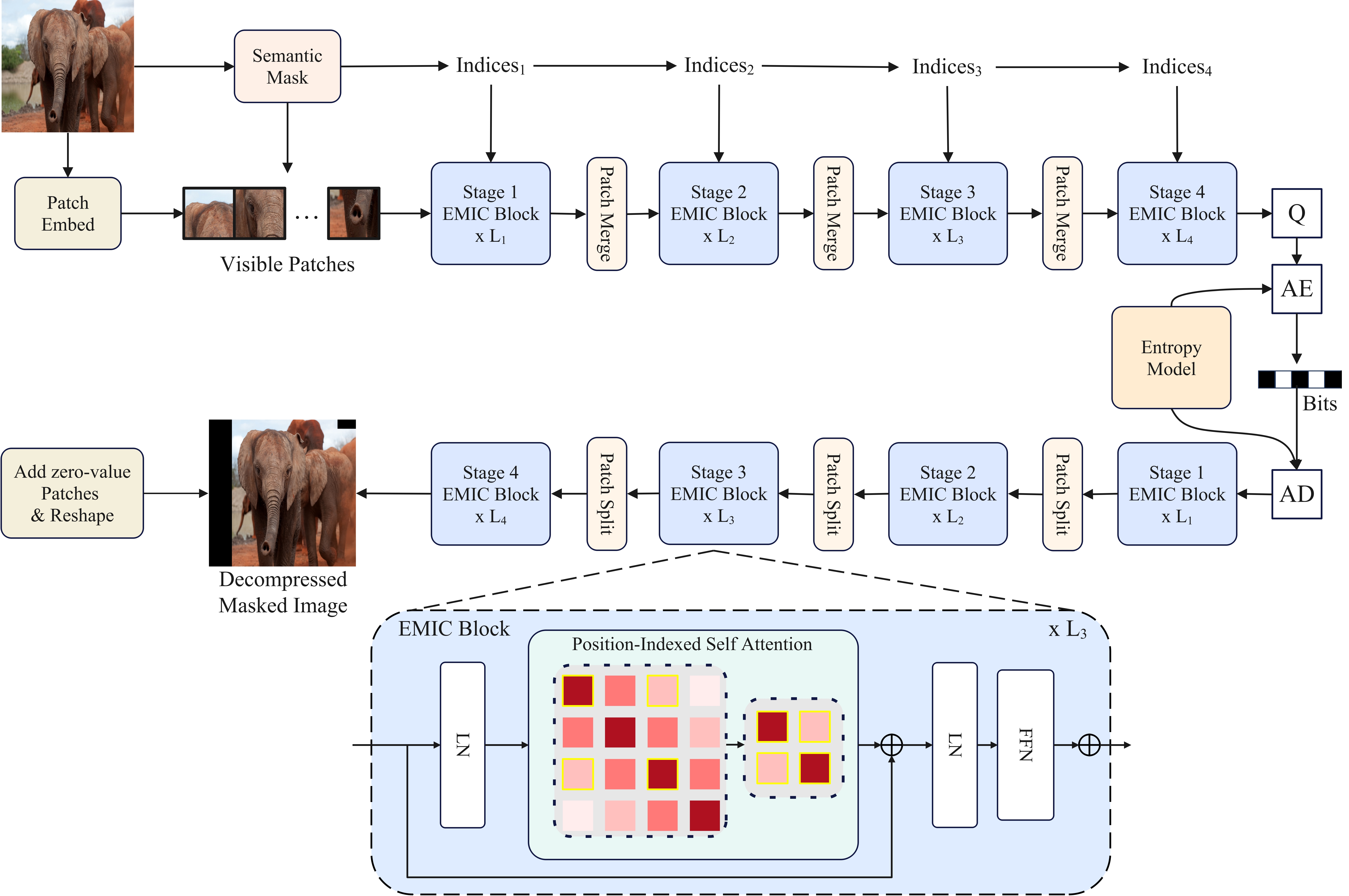}
\caption{The framework of the masked image compression network. Q denotes quantization, AE denotes arithmetic encoding, AD denotes arithmetic decoding, and $L_i$ represents the number of EMIC blocks in the i-th stage (In this paper, $L_1$ to $L_4$ are 2, 2, 6 and 2, respectively). The upper part is the encoding process, while the lower part is the decoding process.}
\label{fig: main structure}
\end{figure*}

\section{Related Work}
\subsection{High-level Vision Tasks}
High-level vision tasks in computer vision are those involve understanding and interpreting the content of images at a more abstract or semantic level. Among them, two common tasks are object detection \cite{Girshick2015, Jocher2023a}, and instance segmentation \cite{He2017}. Corresponding neural network models are widely used across various industries to reduce the need for manual image inspection. As numerous images are ultimately processed by these high-level vision models, image compression tailored to such models has drawn increasing attention from researchers. The compression framework proposed in this paper uses detection head of Mask R-CNN \cite{He2017} and YOLOv8 \cite{Jocher2023a} to evaluate the detection performance on compressed images. For segmentation performance on compressed images, we use segmentation head of Mask R-CNN and YOLOv8.

\subsection{Learned Image Compression}
Current learning-based image compression models \cite{He2022,Zou2022,Liu2023} achieve remarkable compression performance by continuously optimizing three main components: the encoder, the decoder and the entropy model. These models have already surpassed current compression standards (e.g., BPG \cite{Bellard2015}, VVC \cite{Bross2021}) in metrics such as PSNR and SSIM. However, these methods are optimized for human visual perception and do not meet the compression requirements for the downstream tasks.

\citeauthor{Feng2022} \cite{Feng2022} proposed omnipotent features compression. They designed an information filtering module to filter out information that is weakly related to downstream tasks. Nevertheless, it still needs fine-tuning the backbone tail to support different tasks. \citeauthor{Chen2023} \cite{Chen2023} proposed a feature compression method for multi-task machine vision. However, with a single training session, it only supports the corresponding backbone during inference. Additionally, both of the aforementioned feature compression methods do not provide decompressed images, making them unsuitable for scenarios where manual verification of targets is required.

\citeauthor{Li2025} \cite{Li2025} proposed an adapter-based tuning framework aimed at optimizing image compression for both machine and human vision. To accommodate different downstream tasks, the method involves fine-tuning the adapter. Notably, the adapter parameters are activated exclusively for machine vision, implying that distinct forward passes are required to generate images tailored for human and machine vision, respectively. Another kind of methods that can simultaneously meet the needs of both machine vision and human vision is semantically structured image compression \cite{Sun2020,Feng2023}. However, the encoder and decoder in existing methods still process information that will be discarded, leading to a waste of computational resources.

\subsection{Masked Image Modeling}
The MAE \cite{He2022a} based on vision transformer (ViT) has led the trend in masked image modeling. The main components of MAE are an encoder and a decoder. The input image is first randomly masked at the patch level, and only visible tokens will be fed into the encoder. For decoding, the encoded features are combined with the masked tokens as the decoder's input to reconstruct the image. Inspired by above process, considering image compression for high-level vision tasks, we can reduce the image size and FLOPs required by the compression model by masking patches that are weakly related to object information during encoding. In this paper, we propose a hierarchical transformer image compression network suitable for masked input, which performs compression and decompression only on the visible image patches.

\section{Proposed Method}
The framework of our masked image compression network is shown in Fig. \ref{fig: main structure}.  First, the embedded images are masked based on the semantic map, retaining only the patches required for the downstream task as visible patches. These visible patches are then encoded by the encoder network. The output of encoder is quantized and subsequently compressed using an arithmetic encoder to generate the bitstream. For decompression, the arithmetic decoder reconstructs the compressed representation. It is important to note that the data distribution required for entropy coding (arithmetic encoding and decoding) is derived from the entropy model. The decompressed representation is processed by the decoder network to recover the visible patches. Finally, use zero-value patches to fill in the missing parts and reshape the patch sequence into an image. In this section, we provide a detailed discussion of the proposed framework, focusing on the encoding and decoding process, as well as the entropy model and loss function.

\subsection{The Encoding \& Decoding process}
\begin{figure}[h]
\centering
\includegraphics[width=0.95\columnwidth]{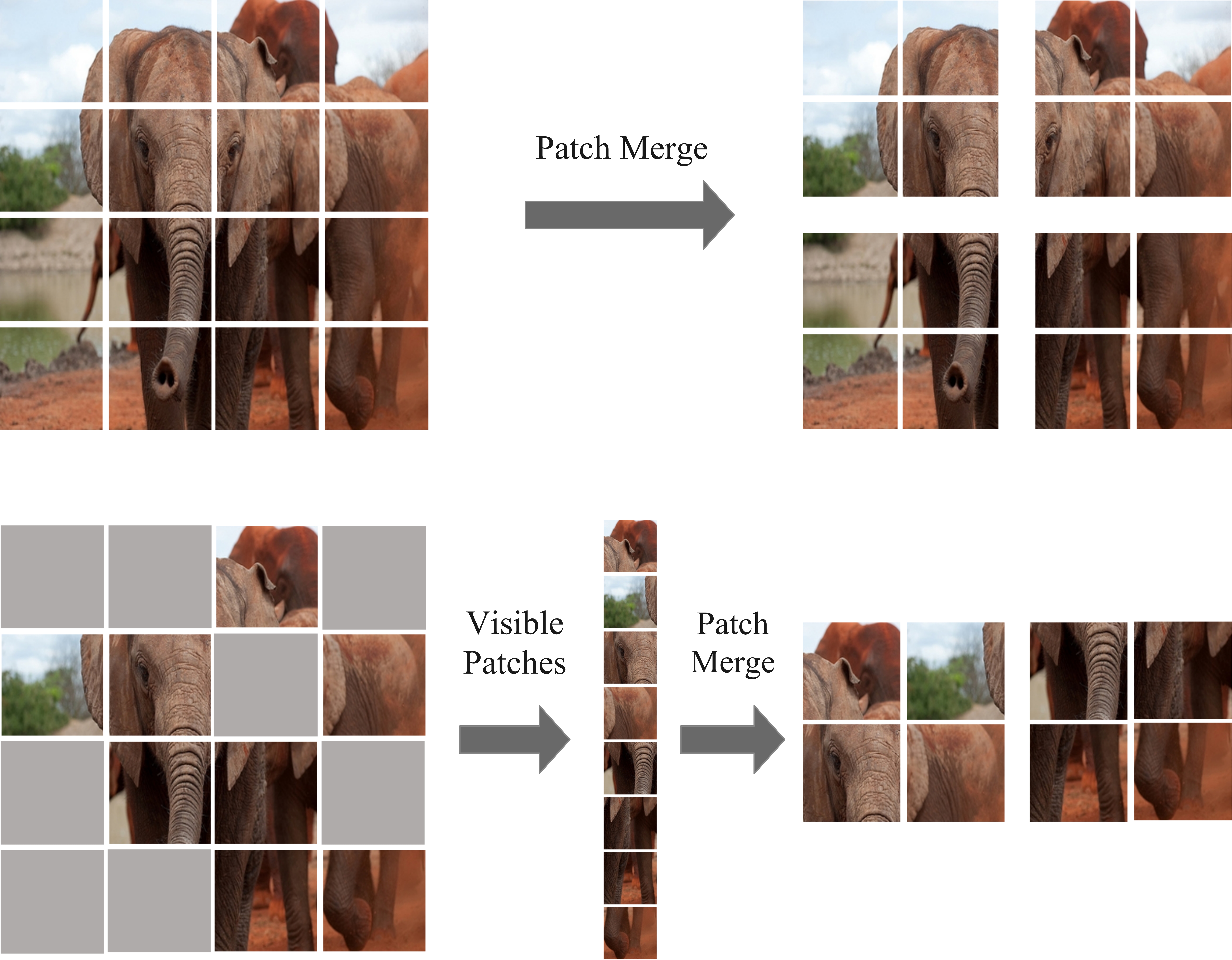}
\caption{Illustration of block merging in the normal case (upper part) and the masked image case (lower part).}
\label{fig: patch merge}
\end{figure}

Since the proposal of MAE \cite{He2022a}, transformer-based structures \cite{Xie2022, Wang2023} have been the primary choice for the encoder and decoder in MIM tasks. When applying masked image modeling to compression tasks, we still consider using transformer-based methods. However, several key issues need to be resolved. First, the image compression model involves dimensionality reduction, which transforms the input image into a more compact feature. The common hierarchical transformer structure \cite{Liu2021,fan2024rmt} achieves dimensionality reduction through a patch merging process. However, as shown in Fig. \ref{fig: patch merge}, in the context of masked image modeling, patch merging affects the relative positions of patches (patches that are far apart in the original image may be merged into one). Therefore, referring to Hiera \cite{Ryali2023}, we distinguish between mask units and attention units during the compression process.

Secondly, images have spatial redundancy, with greater redundancy between closer patches. Therefore, during the compression process, we want the self-attention mechanism to better utilize the information from nearby patches and effectively aggregate their information.  RMT\cite{fan2024rmt} combines vision transformers with mechanisms from retentive networks \cite{Sun2023}, and introduces Manhattan self-attention to embed bidirectional and two-dimensional spatial decay into the attention calculation. This explicit spatial prior better captures positional relationships between tokens. Inspired by this, we propose a position-indexed attention mechanism that utilizes the indices of visible patches to implement Manhattan self-attention, focusing specifically on interactions within the visible patches.

\subsubsection{Mask Unit \& Attention Unit}
\begin{figure}[h]
\centering
\includegraphics[width=0.9\columnwidth]{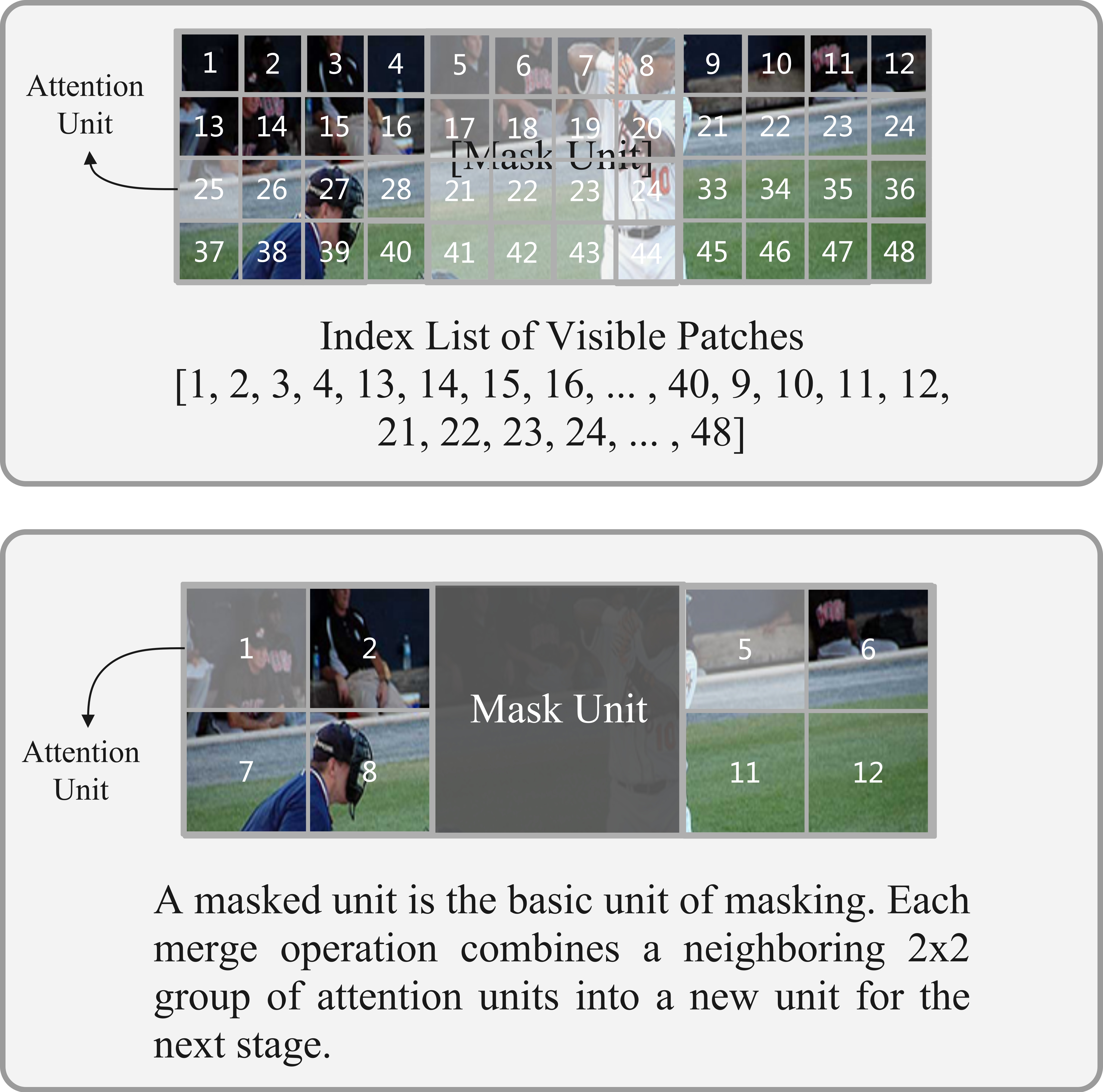}
\caption{The attention unit and mask unit.}
\label{fig: mask unit}
\end{figure}

We define the basic unit of the mask operation as a 16$\times$16 patch. Initially, an attention unit is defined as a 2$\times$2 patch. At the beginning, each mask unit corresponds to 64 attention units. In the first three stages of the encoder, a patch merge operation is performed after each stage (Fig. \ref{fig: mask unit}). Following the swin transformer \cite{Liu2021}, the patch merge operation concatenates the features of each group of 2$\times$2 neighboring attention unit (with C channels) and applies a linear layer on the 4C-dimensional concatenated features. This way, the area corresponding to each attention unit in the original image is enlarged by four times, and the output dimension is set to 2C. Finally, after three times patch merging, the space size of an attention unit is equal to that of a mask unit. 

\subsubsection{Position-Indexed Self-Attention \& Decomposed Attention}
\begin{figure}[h]
\centering
\includegraphics[width=0.8\columnwidth]{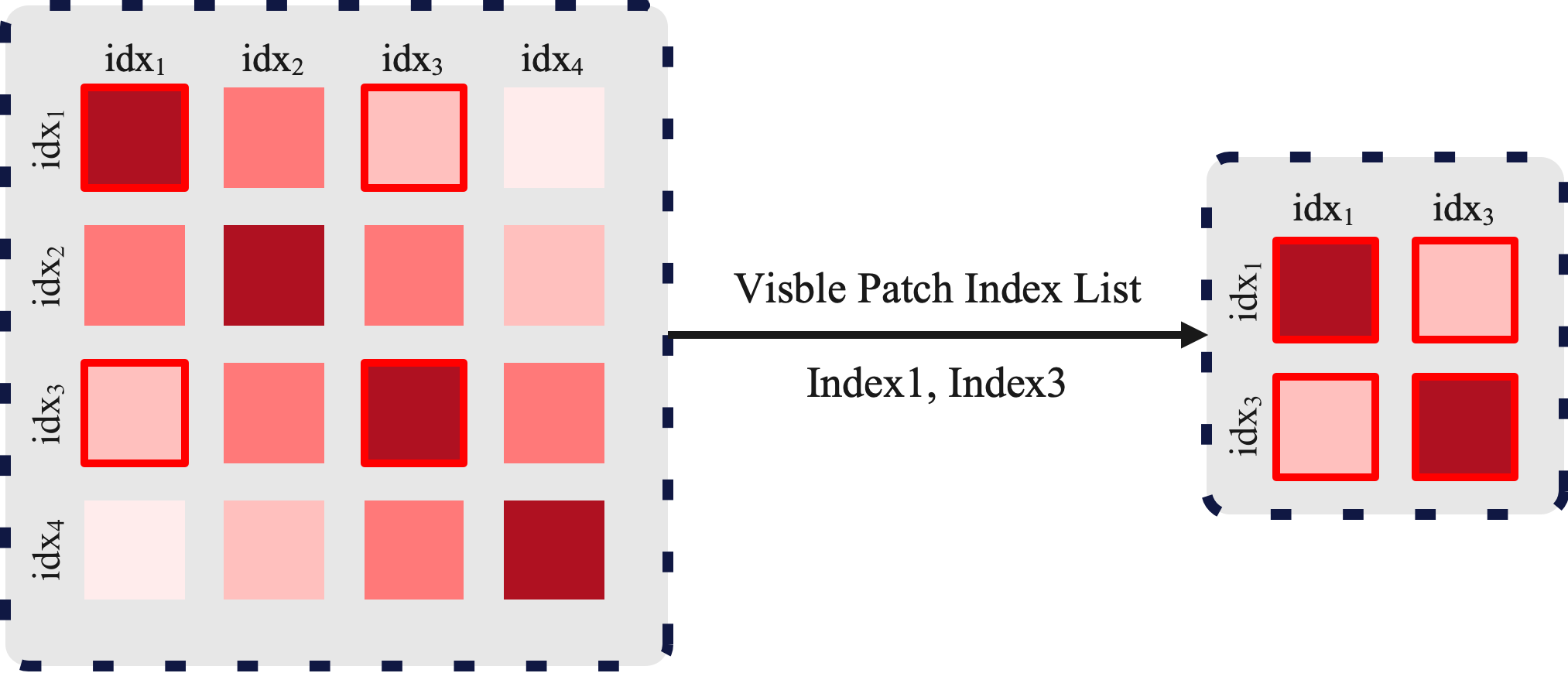}
\caption{Sample the current Manhattan distance decay matrix based on the position indices of the visible blocks.}
\label{fig: manhattan matrix}
\end{figure}

After patch embedding and semantic masking, the input image is transformed into a sequence containing only the visible patches. Consequently, in the encoding stage, unlike RMT \cite{fan2024rmt}, which calculates the Manhattan distance between all patches in the image, our approach only computes the Manhattan distances between the visible patches. Fig. \ref{fig: manhattan matrix} illustrates the Manhattan distance matrix for a total patch length of 4, alongside the corresponding Manhattan distance matrix for visible patches with a length of 2. The resulting Manhattan matrix serves as a spatial prior to guide the computation of self-attention.

It can be seen that the calculation process of Manhattan self-attention at each encoding stage requires the position indices of the current patches, and we name the whole process as position-indexed self-attention (PISA). As shown in Fig. \ref{fig: mask unit}, the index list records the position indices of each attention unit (visible patches). The position indices are arranged in raster order. Since the first three stages involve patch merging operations, the position indices of the current stage also need to be transformed to obtain the position indices for the next stage. Assume the Index List for stage $k$ is $Indices_k = [idx_1, idx_2, \ldots, idx_n]$, the ratio of the width of the input image to the size of the current attention unit is $r_1$, and the ratio of the mask unit size to the attention unit size is $r_2$. The calculation of $Indices_{k+1}$ is as follows:

\begin{algorithm}
\caption{Calculation of $Indices_{k+1}$}
\label{algorithm:1}
\begin{algorithmic}[1]
    \STATE \textbf{Input}: $Indices_k$, Ratio $r_1$, Ratio $r_2$
    \STATE $n \gets \frac{k}{r_2^2}$
    \STATE $Indices_k \gets Indices_k.reshape(n, r_2, r_2)[:, ::2, ::2]$
    \STATE $Indices_k.reshape(-1)$
    \STATE $Indices_{k+1} \gets \frac{Indices_k}{r_1 \times 2} \times \frac{r_1}{2}+ \frac{Indices_k \bmod r_1}{2}$
\end{algorithmic}
\end{algorithm}

In the early stages, the large number of visible attention units results in significant computational overhead for self-attention as it tries to capture global information. RMT addressed such issue by calculating attention scores separately for the horizontal and vertical directions in the image. Since our input is a sequence of visible attention units, decomposition along the horizontal and vertical directions is not feasible. Instead, we first compute the attention at the mask unit level, and then compute the attention within each mask unit. The calculation of decomposed position-indexed self-attention (DPISA) is as follows:

\begin{equation}
\begin{split}
    D_{mn}^{MU} &= \gamma^{|x_m-x_n| + |y_m-y_n|}, \\ 
    D_{kt}^{AU} &= \gamma^{|u_k-u_t| + |v_k-v_t|}, \\
    Attn_{MU} &= Softmax(Q_{MU}K_{MU}^{T}) \odot D_{mn}^{MU}, \\ 
    Attn_{AU} &= Softmax(Q_{AU}K_{AU}^{T}) \odot D_{kt}^{AU}, \\ 
    DPISA(x) &= Attn_{AU}(Attn_{MU}V)^T
\end{split}
\end{equation}

Where the coordinates $(x, y)$ represent the position of the mask unit within the entire image, $(u, v)$ denote the coordinates of the attention unit within a mask unit, $D_{mn}^{MU}$ is the spatial decay matrix at mask unit level, and $D_{kt}^{AU}$ is the spatial decay matrix within a mask unit (i.e., at attention unit level). Both spatial decay matrices are calculated based on the position indices.

The decoding stage implements the reverse process of the encoding stage. Unlike MAE, our approach computes only on the visible patches during decoding. The Patch Split operation consists of a linear layer and a pixel shuffle operation, which are used to restore the image resolution. The decoding stage continues to use position-indexed self-attention to aggregate information, and different stages of the decoder also require transformations of the position indices, with the calculation method as Algorithm ~\ref{algorithm:2}.

\begin{algorithm}
\caption{Calculation of $Indices_{k+1}$}
\label{algorithm:2}
\begin{algorithmic}[1]
    \STATE \textbf{Input}: $Indices_k$, Ratio $r_1$, Ratio $r_2$
    \STATE $n \gets r_2 \div 2$
    \STATE $Indices_{\text{tmp}} \gets \left( \frac{Indices_k}{r_1} \times 2 \times 2 \times r_1 \right) + \left( Indices_k \bmod r_1 \times 2 \right)$
    \STATE $bias \gets \text{tensor}((0, 1, 2 \times r_1, 2 \times r_1 + 1))$
    \STATE $Indices_{k+1} \gets Indices_{\text{tmp}}[:, :, \text{None}] + bias[\text{None}, \text{None}, :]$
    \STATE $Indices_{k+1} \gets Indices_{k+1}.reshape(n, 2, 2)$
    \STATE $Indices_{k+1} \gets Indices_{k+1}.permute(1, 0, 2)$
    \STATE $Indices_{k+1} \gets Indices_{k+1}.reshape(-1)$
\end{algorithmic}
\end{algorithm}

After decoding, zero-valued patches are appended to the decompressed visible patches produced by the decoder, and the complete sequence of patches is reshaped into a 2D structure to reconstruct the masked image.

\subsection{The Entropy Model}

The entropy model in neural network image compression is used to estimate the probability distribution of latent representations, and the most advanced methods all \cite{Balle2018,Minnen2018,Cheng2020,He2021,Liu2023} utilize hyperpriors to model the distribution more accurately. The coding process of hyperpriors is commonly implemented using convolution \cite{Cheng2020,Minnen2018}. However, similar to the challenges faced by image encoders and decoders, the spatial locations of each element in the latent representations are not adjacent in the image due to the mask operation, making it difficult for the convolution to capture appropriate local information. Though Entroformer \cite{Qian2022} proposed a transformer based entropy model, it still uses convolution for downscale and upscale in the entropy model.

\begin{figure}[h]
\centering
\includegraphics[width=\columnwidth]{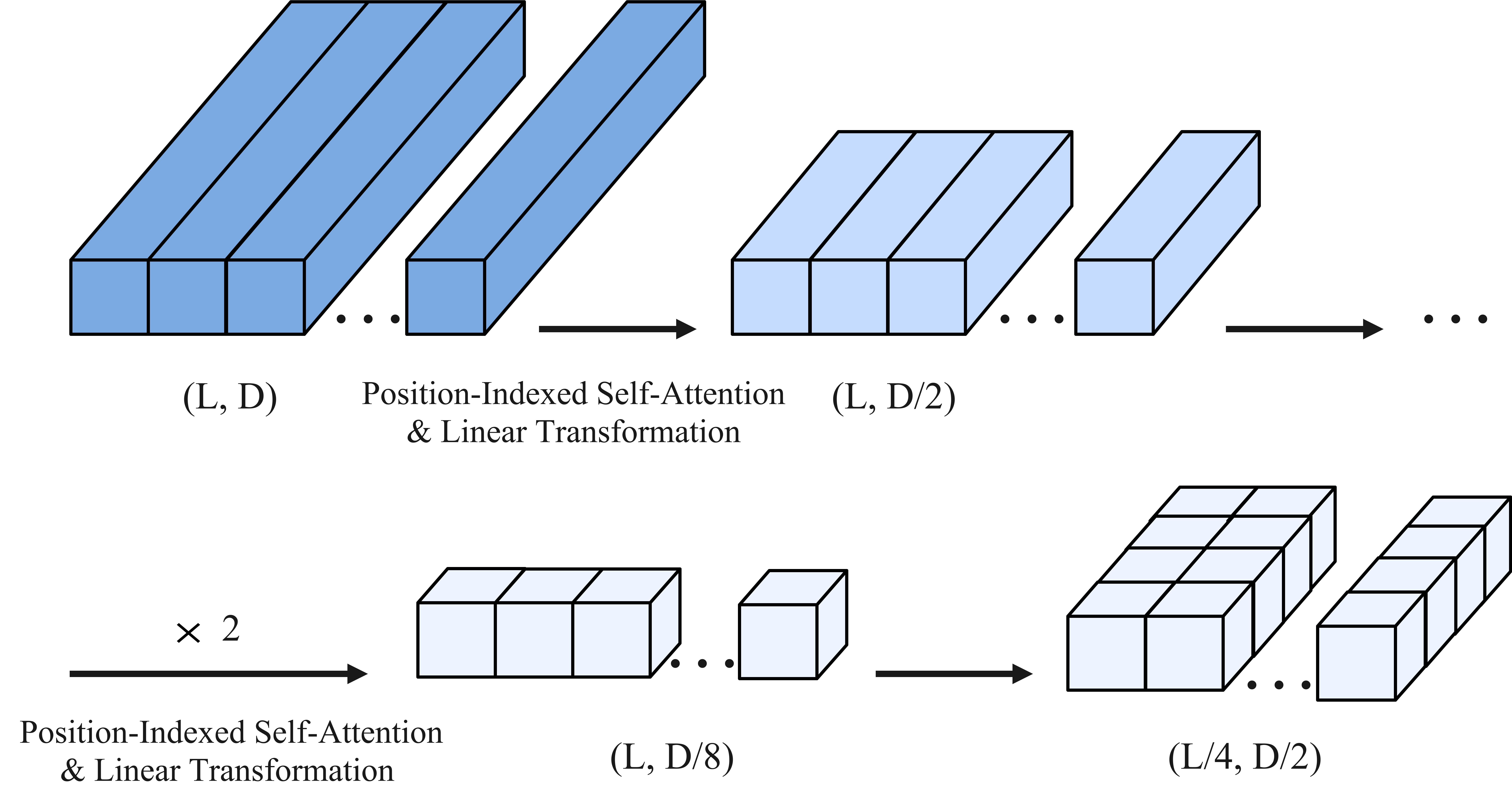}
\caption{The encoding process of the entropy model.}
\label{fig: entropy model}
\end{figure}

\begin{figure*}[t]
\centering
\includegraphics[width=\textwidth]{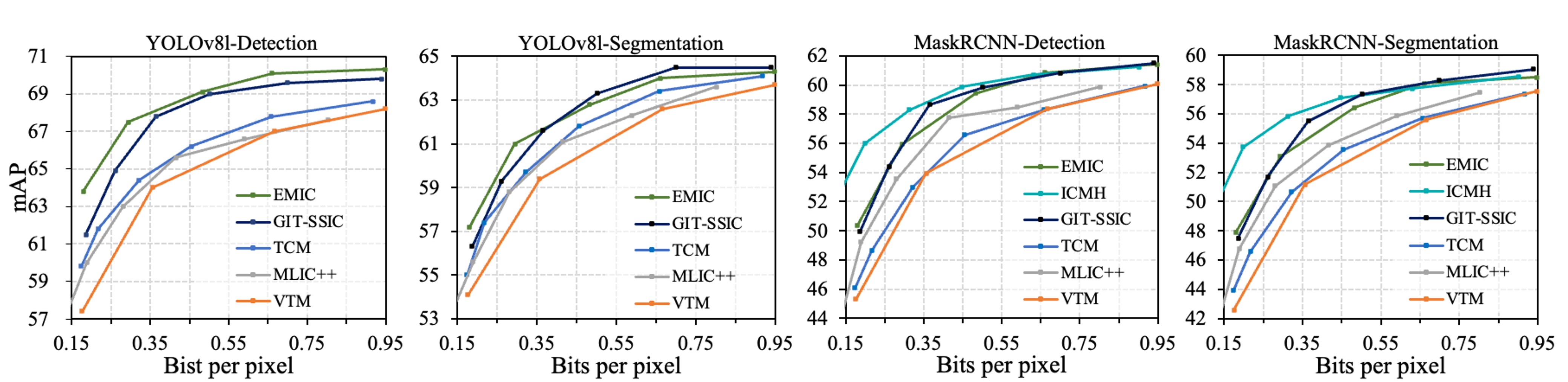}
\caption{Rate-Accuracy performance evaluation of object detection and instance segmentation tasks on the COCO dataset. We use Mask R-CNN (based on ResNeXt-101 and FPN) and YOLOv8l as the high-level vision model.}
\label{fig: RA curve}
\end{figure*}

We propose a method entirely based on attention and linear transformation. As shown in Fig. \ref{fig: entropy model}, since the positional relationship between patches does not change when using linear transformation to reduce channel dimensions, the hyper encoder can use position-indexed self-attention to leverage the spatial information between patches just like the encoder. Finally, every four adjacent patches are concatenated along the channel dimension. Assuming the latent representation has the shape (L, D), the shape of the hyper encoder output will be (L/4, D/2). This process successfully aggregates the information of the latent representation and reduces its length to obtain the hyper latent representation. For the decoder, it first splits the hyper latents by channels, restoring them to a size of (L, D/8). Then, it uses position-indexed self-attention and linear transformation to increase the channels. To utilize channel redundancy, we follow the work of \citeauthor{Minnen2020} \cite{Minnen2020} and use a channel-wise autoregressive process. Therefore, the decoder has one more layer than the encoder, resulting in features of (L, D*2).

\subsection{The Loss Function}
Following the most commonly used approach, the loss function is defined as a rate-distortion loss:
\begin{equation}
\begin{split}
    L &= \underbrace{R(\hat{y}) + R(\hat{z})}_{\text{BPP loss}} + \lambda \cdot \underbrace{D(x, \hat{x} \mid mask)}_{\text{Distortion loss}} \\
    &= \mathbb{E}[-\log_2(p_{\hat{y}|\hat{z}}(\hat{y}|\hat{z}))] + \mathbb{E}[-\log_2(p_{\hat{z}|\phi}(\hat{z}|\phi))]  \\
    &\quad + \lambda \cdot D(x, \hat{x} \mid mask)
\end{split}
\end{equation}
Where $x$ is the input image, $R(\hat{y})$ is the rate of the latent representation, $R(\hat{z})$ is the hyper latent representation and $D(x, \hat{x})$ is the distortion (MSE loss) between input image and decompressed image. It is important to note that when calculating the distortion loss, we only need to consider the distortion of the visible patches:
\begin{equation}
    D(x, \hat{x} | mask) = \frac{\sum_{m_{ij} \neq 0}(x_{ij} - \hat{x}_{ij})^2}{\sum_{m_{ij} \neq 0} m_{ij}} \quad \text{where} \quad m_{ij} \in mask
\end{equation}
Similarly, when computing the rate loss, the total number of pixels used for division should be the number of pixels in the visible patches:
\begin{equation}
    \text{BPP Loss} = \frac{R(\hat{y}) + R(\hat{z})}{\sum_{m_{ij} \neq 0} m_{ij}} \quad \text{where} \quad m_{ij} \in mask
\end{equation}

\section{Experiments}

\subsection{Dataset}
The COCO\cite{Lin2014} and Kodak24\cite{kodak1993} datasets are used for the experiments.

\textbf{COCO.} To evaluate the performance of compressed images in object detection and instance segmentation tasks, we use the widely adopted COCO dataset for model training and evaluation. It contains 118,287 training images and 5,000 validation images, with over 500,000 annotated object instances for detection and segmentation. We randomly sample 500 images along with their corresponding labels from the validation set for testing purposes.

\textbf{Kodak24.} We hope that the compressed images obtained through our model not only perform well in compression rate but also have good compression efficiency. The Kodak24 dataset contains 24 color images of different scenes and subjects and is widely used for testing image compression algorithms. Therefore, we use Kodak24 to evaluate the compression efficiency of different models.

\subsection{Implementation Details}
We train the compression model on the COCO dataset using the RTX 4090 GPU, randomly cropping the input images to 256$\times$256 at each iteration. To obtain models with different compression rates, we set the $\lambda$ values in the rate-distortion loss to \{0.004, 0.01, 0.025, 0.05, 0.1\}. All models are trained for 150 epochs using the Adam optimizer \cite{Kingma2014} with a batch size of 8. We use the ReduceLROnPlateau learning rate scheduler in PyTorch \cite{Paszke2019} to control the learning rate. The initial learning rate is set to 1 $\times$ $10^{-4}$, with patience set to 10 and factor set to 0.3. To ensure the compression model adapts to diverse masking scenarios, we generate random group masks during training following the same method used in GIT-SSIC.

We evaluate the performance of the compression model on high-level vision tasks using object detection and instance segmentation. For object detection, we utilize YOLOv8L-detection and MaskRCNN-detection (with an X101-FPN backbone). For instance segmentation, we employ YOLOv8L-segment and MaskRCNN-segment (also with an X101-FPN backbone). YOLOv8 is implemented using Ultralytics \cite{Jocher2023a}, while MaskRCNN is implemented using Detectron2 \cite{Wu2019}. All models are evaluated using images with a resolution of 512 $\times$ 512.

\subsection{Results}
\label{sec: results}
Fig. \ref{fig: RA curve} illustrates the Rate-Accuracy (R-A) curves, comparing our method with existing learned image compression models, including TCM \cite{Liu2023}, MLIC++\cite{Jiang2023}, GIT-SSIC \cite{Feng2023}, and Adapt-ICMH \cite{Li2025}, as well as the advanced traditional handcrafted compression standard VVC (VTM 22.2). Among these, TCM, MLIC++ and VTM are designed exclusively for human perception, while GIT-SSIC and ICMH also take machine vision into account. Similar to the Rate-Distortion (R-D) curve used for evaluating traditional image compression performance, the R-A curve is used to evaluate the performance of compression models for high-level vision tasks. The vertical axis of the curve represents mAP@0.5 (mean Average Precision at an IoU threshold of 0.5), a commonly used metric for object detection and instance segmentation, while the horizontal axis represents the bits per pixel (BPP) of the image.

As shown in Fig. \ref{fig: RA curve}, Adapt-ICMH achieves higher detection accuracy at high compression rates (i.e., low bpp). However, it has the following limitations compared to semantic mask-based compression methods:
\begin{enumerate}
    \item The reconstructed image cannot simultaneously satisfy the requirements of both human and machine vision. Specifically, the image for human observation is produced through a forward pass of the base codec, while the version used for machine processing requires an additional forward pass with the adapter applied. As a result, when both manual verification and downstream high-level vision tasks are required, Adapt-ICMH incurs double the encoding and decoding time, and necessitates storing two separate sets of intermediate results.
    \item For different machine vision tasks, the adapter must be fine-tuned individually. In contrast, semantic mask-based compression methods require only a single training process to generalize across multiple downstream tasks.
\end{enumerate}

Among the compression models compared, both our model and GIT-SSIC use semantic masking for image compression. In the experiments, to eliminate confounding factors, we generate semantic masks based on ground-truth detection labels from the COCO dataset. In these masks, pixels within all target bounding boxes are assigned a value of 1, while all other regions are set to 0. As shown in Fig. \ref{fig: RA curve}, our model achieves performance comparable to that of GIT-SSIC across various downstream tasks. Furthermore, in comparison with MLIC++, TCM, and VTM—which are designed exclusively for human visual perception, EMIC exhibits a clear advantage in compression scenarios targeting downstream vision tasks.

In image compression tasks, encoding time, decoding time, and computational complexity are also important performance metrics. To this end, we compare the FLOPs, total number of parameters, encoding time, and decoding time on the Kodak dataset across both semantic-aware compression models and image compression models that require only a single forward pass. The two semantic mask-based models, EMIC and GIT-SSIC, adopt the same mask generation strategy. Since the Kodak dataset is not specifically designed for vision tasks and lacks semantic annotations, we generate synthetic semantic masks by randomly selecting 1 to 10 “object” regions per image and treating the remaining areas as masked regions. The total area of non-object regions is determined by a fixed masking ratio. We evaluate model performance under masking ratios of 20\%, 40\%, and 60\%, and the final result is obtained by averaging over these settings. VTM, which is a hand-crafted, CPU-based codec with inferior performance compared to learned image compression methods (see Fig.~\ref{fig: RA curve}), is excluded from this comparison.

As shown in Table~\ref{tab: complexity table}, EMIC demonstrates significantly higher computational efficiency than GIT-SSIC, with its FLOPs reduced to approximately one-third of that required by GIT-SSIC. The results indicate that, while maintaining comparable performance in other aspects, EMIC requires significantly fewer computational resources than GIT-SSIC. Moreover, EMIC achieves the fastest encoding time among all compared models. Although TCM and EMIC exhibit similar overall coding time, TCM lags significantly behind EMIC in compression performance for downstream vision tasks (Fig.~\ref{fig: RA curve}). This improvement in coding efficiency is primarily attributed to our strategy of selectively masking image regions prior to encoding, which enables focused processing of the remaining visible patches during both the encoding and decoding stages. In contrast, due to structural limitations, GIT-SSIC applies masking after encoding and is therefore unable to leverage the mask to reduce computational overhead in earlier stages.

\begin{table}[h]
  \centering
  \caption{ Comparison of model complexity on Kodak.}
  \resizebox{\columnwidth}{!}{
    \begin{tabular}{ccccc}
    \hline
    Model & FLOPs & Parameters & Encode & Decode  \\
    & G& M & ms & ms  \\
    \hline
    \textbf{EMIC} & \textbf{63.32} & \textbf{20.98} & \textbf{160}  & \textbf{181}  \\
    GIT-SSIC & 193.42 & 23.13 & 240 & 150  \\
    TCM & 211.36 & 44.97 & 177 & 160  \\
    MLIC++ & 329.45 & 116.48 & 182 & 241  \\
    \hline
    \end{tabular}%
  }
  \label{tab: complexity table}%
\end{table}%

In many cases, the objects detected by high-level vision models still require manual review and verification. This means that compressed images must not only perform well in high-level vision tasks but also preserve high visual quality in object regions to meet human observation needs. To evaluate the visual quality of images compressed by EMIC, we compare several quality metrics across different compression methods, including PSNR, LPIPS\cite{Zhang2018}, MS-SSIM\cite{Wang2003}, and FID\cite{Heusel2017}. All metrics are computed within the semantic mask-retained regions: the same mask is applied to both the original and compressed images, and the metrics are calculated between these masked versions.

\begin{figure}[h]
\centering
\includegraphics[width=\columnwidth]{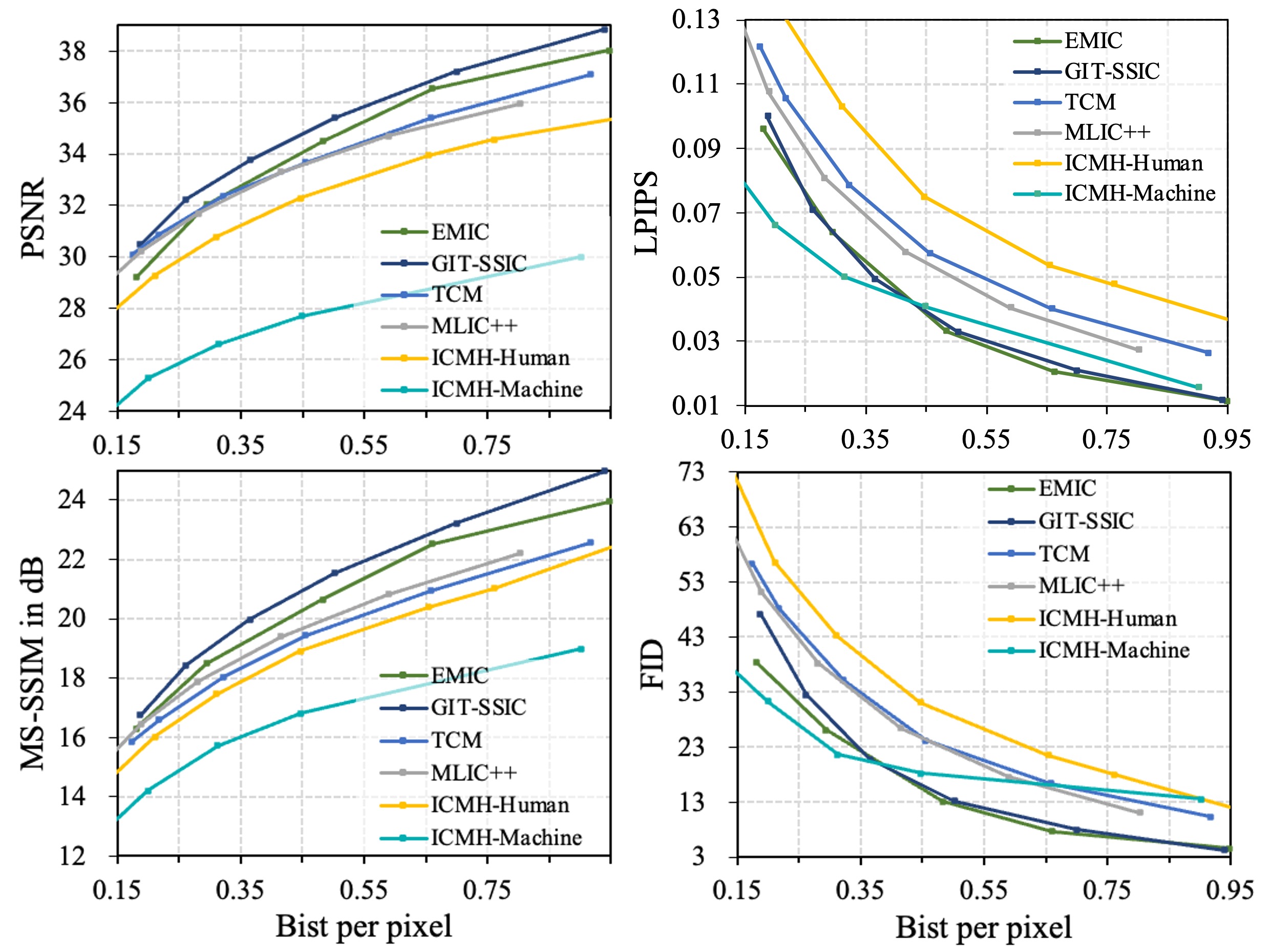}
\caption{Rate-Distortion performance of object regions on the COCO dataset.}
\label{fig: RD curve}
\end{figure}

The comparison results of object-region visual quality are shown in Fig. \ref{fig: RD curve}. Among the compared models, “ICMH-Machine” refers to the Adapt-ICMH model optimized for high-level vision tasks, while “ICMH-Human” refers to a version tailored for human visual perception. It can be seen that EMIC performs very similarly to GIT-SSIC across all quality metrics, indicating that EMIC can achieve a comparable visual quality for human perception. Notably, when ICMH is applied to machine vision tasks (i.e., ICMH-Machine), although it performs well in high-level vision, its PSNR and MS-SSIM scores drop significantly. Even the ICMH-Human version performs worse than both EMIC and GIT-SSIC in terms of all metrics.

Based on Fig. \ref{fig: RA curve}, Fig. \ref{fig: RD curve}, and Table \ref{tab: complexity table}, we can draw the following conclusion: EMIC not only matches or outperforms existing compression models in terms of detection performance and visual quality, but also benefits from a structural advantage. With just a single training process and one forward pass, EMIC can generate images that serve both machine vision models and human inspection. Furthermore, it leverages masking to reduce the computational cost of compression.

\begin{figure}[h]
\centering
\includegraphics[width=\columnwidth]{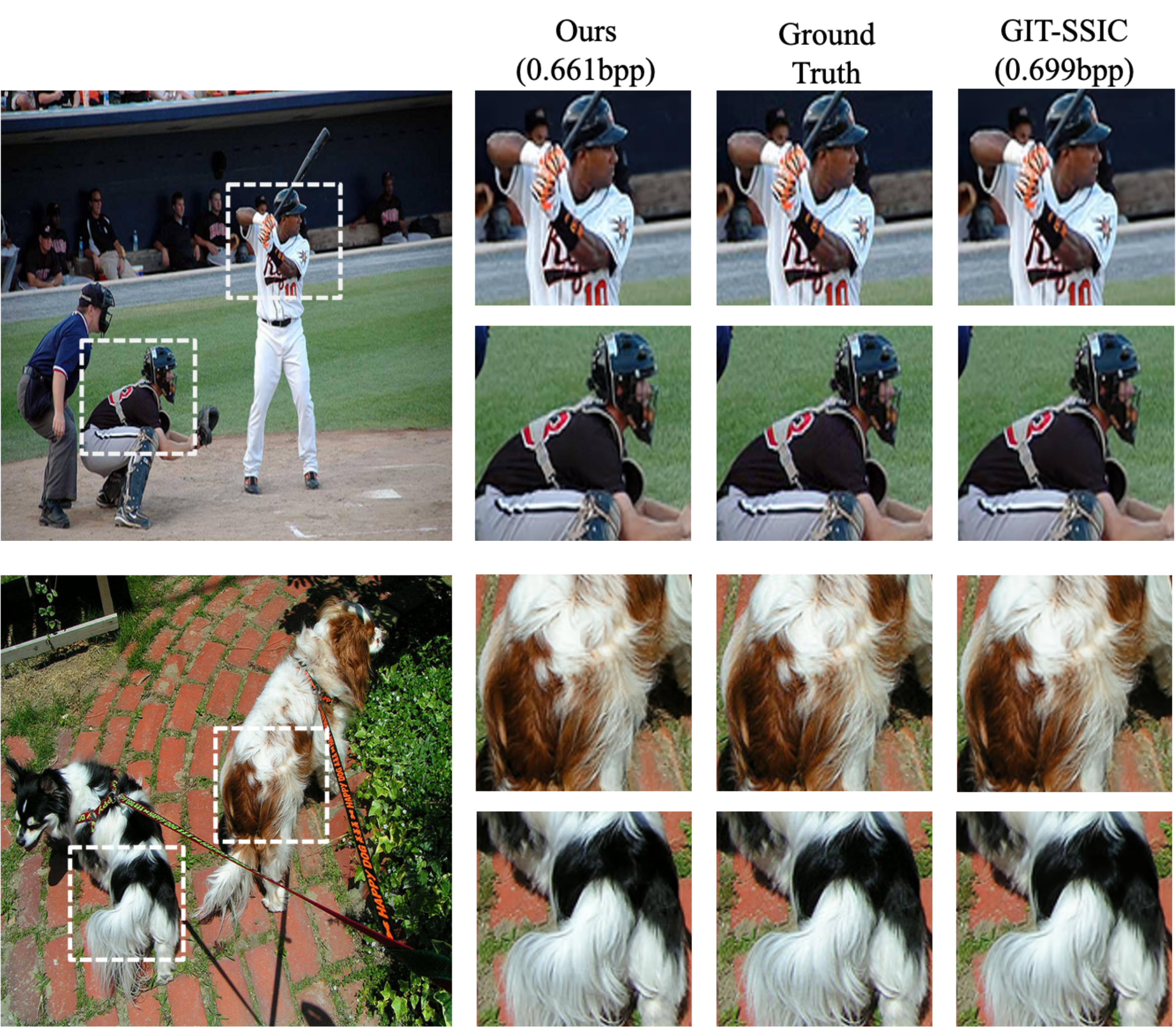}
\caption{Visualization of the object regions in compressed image.}
\label{fig: visualization}
\end{figure}

Fig.~\ref{fig: visualization} shows the object regions in compressed images obtained using our method and GIT-SSIC. At similar bpp levels, both methods produce very similar visual effects, demonstrating that our approach achieves visual quality comparable to GIT-SSIC, while requiring only one-third of its FLOPs.

\section{Ablation Study}
\begin{figure}[h]
\centering
\includegraphics[width=\columnwidth]{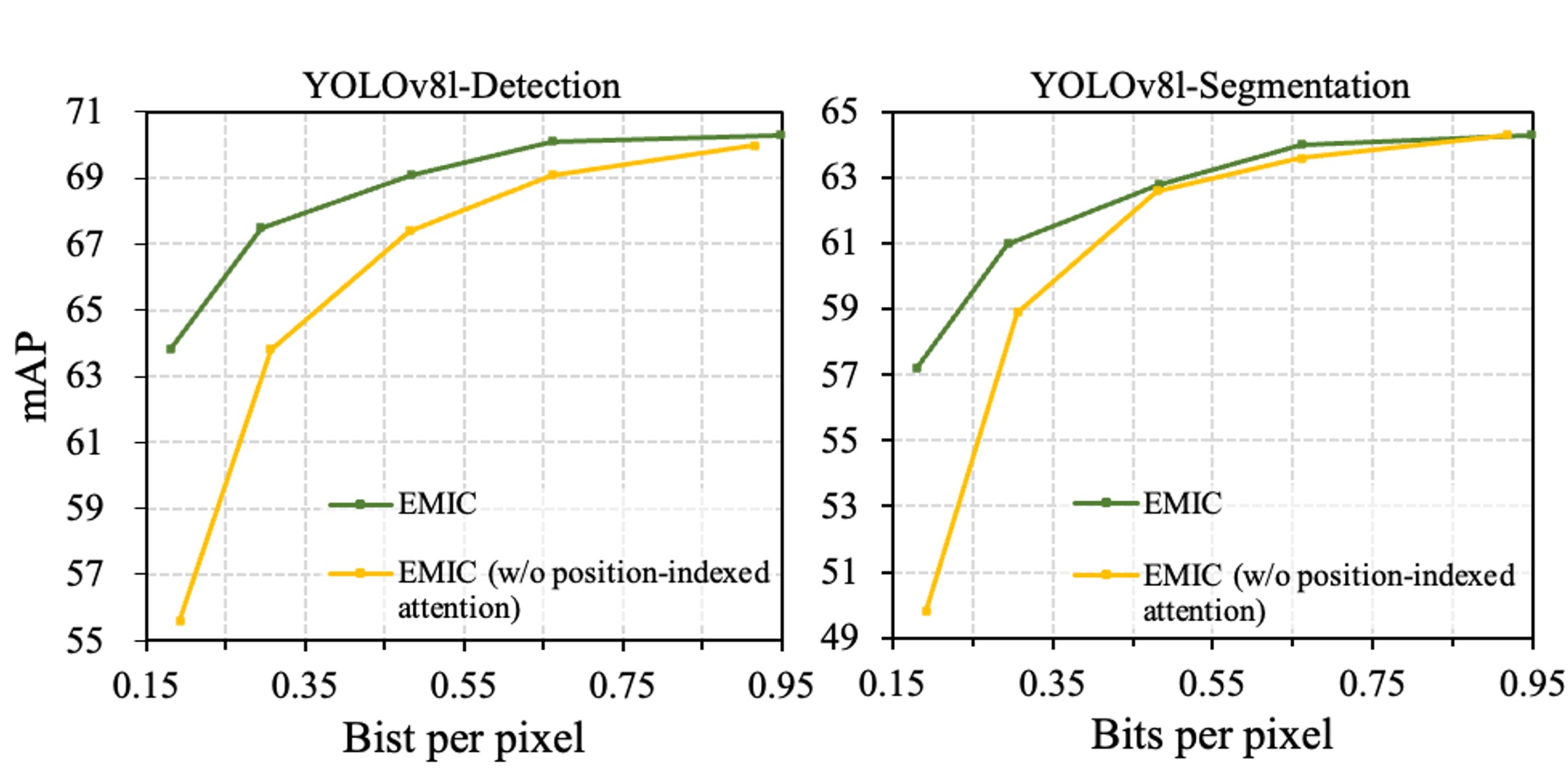}
\caption{Ablation study of the position-indexed attention.}
\label{fig: ablation study}
\end{figure}

To enhance compression performance on the visible patch sequence, we propose a position-indexed attention mechanism. EMIC uses the position indexes of the patches preserved at each stage to compute a Manhattan distance decay matrix, which is then applied to spatially weight the attention matrix. To validate the effectiveness of position-indexed attention, we conduct an ablation study, compressing images from the COCO dataset using both the original model and a variant that excludes position-indexed attention. We then evaluate the decompressed images on object detection and segmentation tasks using YOLOv8-L. As shown in Fig. \ref{fig: ablation study}, the model incorporating position-indexed attention significantly outperforms the one without it, particularly at higher compression rates. This improvement can be attributed to the higher redundancy found in spatially adjacent regions of image data. Position-indexed self-attention provides spatial priors for the unordered patch sequence, strengthening the attention between patches that are spatially close in the original image.

Another key innovation of this work is the use of masking to effectively reduce the computational burden during image compression. To validate its effectiveness, we conduct a detailed statistical analysis of FLOPs, encoding time, and decoding time for both GIT-SSIC and EMIC under varying masking ratios using the Kodak dataset. The mask generation process for different masking ratios follows the same procedure described in Section~\ref{sec: results}. Note that the varying masking ratios are used solely to evaluate the impact of masking on computational load. In practical scenarios, the mask is determined by the semantic content of the image, and the masking ratio is fixed for each instance.

As shown in Table~\ref{tab: mask ratio ablation}, EMIC exhibits a significant reduction in encoding and decoding time as the masking ratio increases, with a corresponding decrease in FLOPs. In contrast, GIT-SSIC’s FLOPs remain unaffected by masking, and its encoding time even increases as the masking ratio grows. This is because GIT-SSIC performs semantic structuring of the bitstream by dividing the image into groups based on connected components. Although the number of objects is close across different masking ratios, lower masking ratios tend to produce larger, more connected object regions, resulting in fewer connected components. Conversely, higher masking ratios lead to smaller and more dispersed targets, thereby increasing the number of connected components. Since GIT-SSIC applies entropy coding sequentially to each connected component, a greater number of components results in longer encoding times. These results demonstrate that EMIC significantly reduces computational load by restricting computations to visible blocks after masking. This validates the effectiveness of semantic masking in enhancing compression efficiency.

\begin{table}[h]
  \centering
  \caption{ Model complexity under different mask ratio.}
  \resizebox{\columnwidth}{!}{
    \begin{tabular}{ccccc}
    \hline
    Model & Mask Ratio &FLOPs & Encode & Decode \\
    & \% & G & ms & ms  \\
    \hline
    \multirow{4}{*}{GIT-SSIC} & 20 & \multirow{4}{*}{193.42} & 216 & 150 \\
    & 40 &  & 237 & 150 \\
    & 60 &  & 268 & 151 \\
    & 80 &  & 306 & 158 \\
    \hline
    \multirow{4}{*}{\textbf{EMIC}} & 20 & 83.93 & 207 & 229 \\
    & 40 & 63.41 & 156 & 182 \\
    & 60 & 42.62 & 117 & 133 \\
    & 80 & 21.83 & 99 & 111 \\
    \hline
    \end{tabular}%
  }
  \label{tab: mask ratio ablation}%
\end{table}%

\section{Conclusion}
In this paper, we propose a novel compression model, EMIC, which is designed to process unordered patch sequences of images after masking. By focusing computations solely on the visible patches, EMIC effectively reduces computational load by avoiding unnecessary processing of irrelevant regions.
Furthermore, we introduce a position-indexed attention mechanism that leverages spatial priors to enhance the effectiveness of patch sequence compression. This mechanism preserves spatial relationships among patches during compression, thereby improving both computational efficiency and reconstruction accuracy.
Experimental results demonstrate that the proposed model significantly improves compression efficiency while maintaining competitive performance in semantic mask-based image compression tasks.

\bibliographystyle{ACM-Reference-Format}
\bibliography{ref}










\end{document}